\documentclass[lettersize,journal]{IEEEtran}
\usepackage{amsmath,amsfonts}
\usepackage{algorithmic}
\usepackage{algorithm}
\usepackage{array}
\usepackage[caption=false,font=normalsize,labelfont=sf,textfont=sf]{subfig}
\usepackage{textcomp}
\usepackage{stfloats}
\usepackage{url}
\usepackage{verbatim}
\usepackage{graphicx}
\usepackage{cite}
\usepackage{tabularx}  
\usepackage{multirow} 
\usepackage{makecell} 
\usepackage{booktabs}

\begin{document}

\title{Large Language models for Time Series Analysis: Techniques, Applications, and Challenges}

\author{Feifei~Shi,
        Xueyan~Yin,
        Kang~Wang,
        Wanyu~Tu,
        Qifu~Sun,
        and~Huansheng~Ning
\thanks{Feifei Shi, Xueyan Yin, Kang Wang, Wanyu Tu, Qifu Sun, and Huansheng Ning are with the School of Computer and Communication Engineering, University of Science and Technology Beijing, Beijing, China, 100083. (Corresponding email: ninghuansheng@ustb.edu.cn)}
\thanks{Manuscript received April 19, 2005; revised August 26, 2015.}}

\markboth{Journal of \LaTeX\ Class Files,~Vol.~14, No.~8, August~2021}%
{Shell \MakeLowercase{\textit{et al.}}: A Sample Article Using IEEEtran.cls for IEEE Journals}


\maketitle

\begin{abstract}

Time series analysis is pivotal in domains like financial forecasting and biomedical monitoring, yet traditional methods are constrained by limited nonlinear feature representation and long-term dependency capture. The emergence of Large Language Models (LLMs) offers transformative potential by leveraging their cross-modal knowledge integration and inherent attention mechanisms for time series analysis. However, the development of general-purpose LLMs for time series from scratch is still hindered by data diversity, annotation scarcity, and computational requirements. This paper presents a systematic review of pre-trained LLM-driven time series analysis, focusing on enabling techniques, potential applications, and open challenges. First, it establishes an evolutionary roadmap of AI-driven time series analysis, from the early machine learning era, through the emerging LLM-driven paradigm, to the development of native temporal foundation models. Second, it organizes and systematizes the technical landscape of LLM-driven time series analysis from a workflow perspective, covering LLMs' input, optimization, and lightweight stages. Finally, it critically examines novel real-world applications and highlights key open challenges that can guide future research and innovation. The work not only provides valuable insights into current advances but also outlines promising directions for future development. It serves as a foundational reference for both academic and industrial researchers, paving the way for the development of more efficient, generalizable, and interpretable systems of LLM-driven time series analysis.

\end{abstract}


\section{Introduction}
\IEEEPARstart{T}{ime} series analysis, as one of the core fields of data science \cite{A2}, plays a crucial role in various scenarios such as financial forecasting, industrial Internet of Things, biomedical monitoring, and climate modeling. Traditional time series analysis methods rely on statistical methods such as AutoRegressive Integrated Moving Average (ARIMA) \cite{A3} and State Space models \cite{A4}, or shallow machine learning algorithms like Support Vector Machine (SVM) and Random Forest (RF), in which the modeling capabilities are limited by the representation of nonlinear features and long-term dependencies. 

In recent years, the Transformer architecture with global attention mechanism has achieved groundbreaking advancements in temporal modeling \cite{A7}. Models like the Temporal Fusion Transformer, utilizing positional encoding and multi-head attention, have demonstrated effective capture of long-range dependencies  \cite{A8}. However, these domain-specific temporal Transformers often require task-specific architectural redesign and suffer from limited parameter scales, hindering their ability to fully exploit latent patterns in massive cross-domain temporal data. Such limitations become particularly pronounced in complex scenarios that require integrated domain knowledge for reasoning, such as coupled analysis of power load forecasting and meteorological factors.

The emergence of Large Language Models (LLMs) \cite{A5} has brought revolutionary opportunities for time series analysis. The general representation capabilities of LLMs with hundreds of billions of parameters have demonstrated strong potential for the integration of cross-modal and cross-domain knowledge. Studies have shown that LLMs can not only process discrete text sequences, but their attention mechanism essentially has the ability to process continuous time series data \cite{A1}. By converting numerical sequences such as sensor readings and stock price fluctuations into token embeddings, LLms can spontaneously learn dynamic patterns in the time dimension, enabling LLMs to combine time series analysis with semantic understanding, such as the simultaneous analysis of physiological signals and patient medical records in medical monitoring \cite{A9}.

However, completely training a general-purpose LLM for time series from scratch is challenging. First, the complexity and diversity of time series data are exceptionally high, with significant variations in sampling rates, noise levels, and periodicity across different domains, requiring that models possess robust cross-domain generalization capabilities. Second, the scarcity of high-quality annotated time series data is particularly acute, as many scenarios rely on domain experts for labeling, which is costly and difficult to scale. Moreover, the computational overhead of training a general-purpose LLM for time series analysis is immense. For example, processing long sequences leads to exponential growth in memory and computing power requirements, while hardware acceleration support for time series operations remains underdeveloped. These factors collectively create significant bottlenecks at the data, algorithmic, and computational levels to develop general LLMs for time series. Therefore, researchers are committed to exploring the potential of pre-trained LLMs in capturing complex dependencies of time series and promoting various applications.

In this survey, we are going to provide an overview of existing methods leveraging pre-trained LLMs for time series analysis. The main contributions are concluded as follows:

\begin{itemize}
    \item Establish a roadmap of AI-driven time series analysis, systematically tracing the evolution from early machine learning era to LLM-driven paradigm, and envisioning future advancements in native temporal foundation models.
    \item Systematize the technical taxonomy of LLMs for time series analysis tasks,  input processing, optimization mechanisms, and lightweight deployment strategies.
    \item Highlight novel real-world applications for LLM-enhanced time series analysis and critically examines open research challenges to guide future innovation.
\end{itemize}

The remainder of this paper is structured as follows. Section II delineates the evolutionary trajectory of AI-driven time series analysis, tracing its progression from the early machine learning era to the current LLM-driven paradigm and finally advancing toward the emerging stage of native temporal foundation models. Section III elaborates on core technical innovations in leveraging LLMs for time series analysis, spanning input, optimization, and lightweight techniques. Section IV showcases state-of-the-art LLMs that advance time series analysis. Section V explores emerging applications empowered by LLM-driven time series methodologies. Section VI discusses critical open issues and challenges in the field, while Section VII provides concluding remarks.

\section{The Evolution Roadmap of AI-driven Time Series Analysis} 

\begin{figure}
\centering
\includegraphics[width=9 cm]{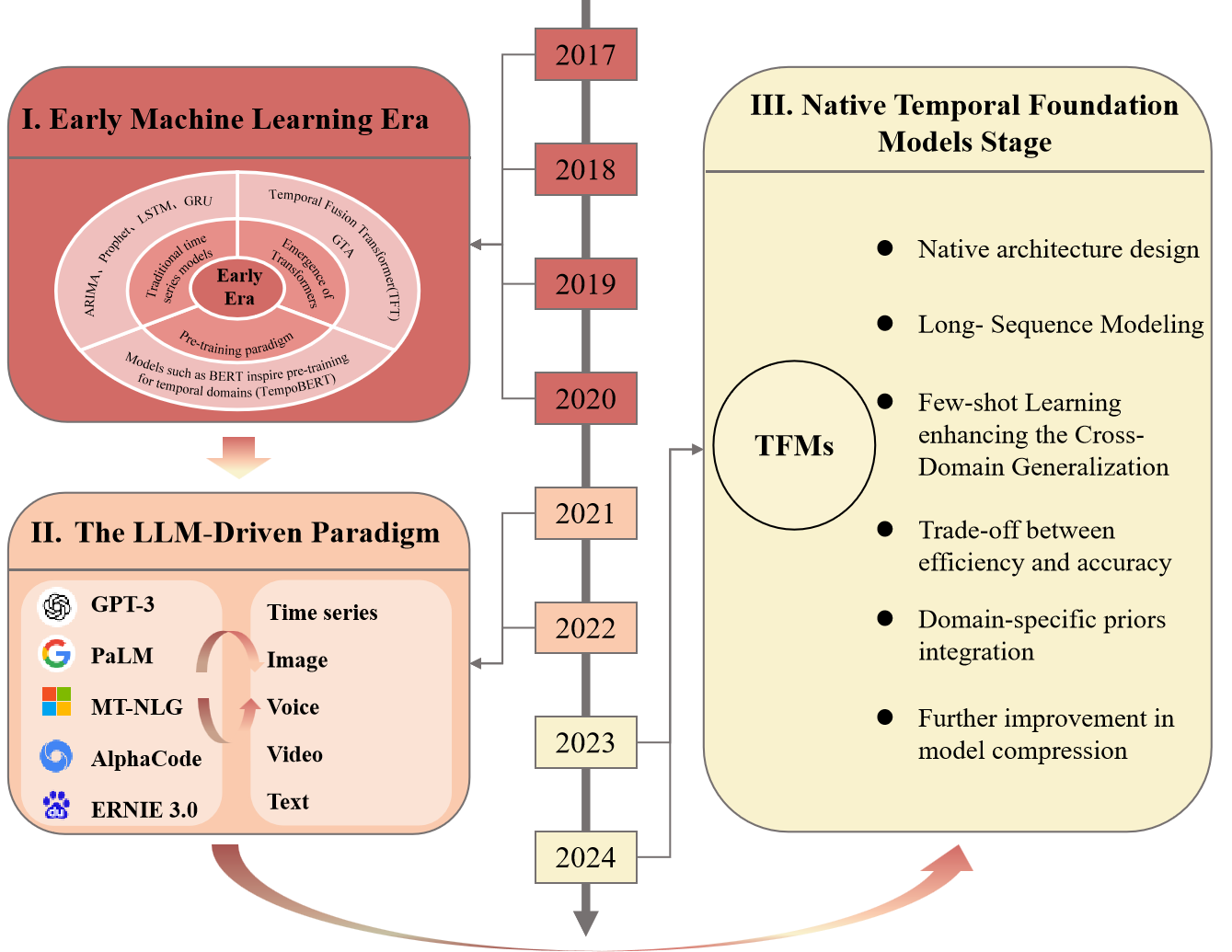}
\caption{The Evolution Roadmap of AI-driven Time Series Analysis.}
\label{fig2}
\end{figure}

In recent years, time series analysis has undergone a paradigm shift, evolving from the early era of machine learning, further accelerated with the rise of LLMs, and developed towards native temporal foundation models. As illustrated in Figure \ref{fig2}, we present a systematic evolution roadmap of AI-driven advances in time series analysis:

\begin{itemize}
    \item \textbf{Early Machine Learning Era}: Characterized by foundational developments in machine learning techniques and early Transformer architecture.
    \item \textbf{The LLM-driven Paradigm}: Marked by pioneering efforts to adapt LLMs for time series analysis, leveraging their capacity for capturing cross-domain dependencies.
    \item \textbf{Native Temporal Foundation Models Stage}: Defined by the purpose-built architectures specifically designed for time series analysis, modling the intrinsic temporal dynamics, and prioritizing domain-specific scalability and interpretability.
\end{itemize}

\subsection{Early Machine Learning Era}

In the early era of time series analysis, methodologies predominantly rely on statistical models and shallow machine learning architectures such as LSTMs (Long Short-Term Memory networks) and GRUs (Gated Recurrent Units). These models have demonstrated efficacy in capturing short-term dependencies and processing low-dimensional time series data. However, their utility was limited by inherent limitations, including fixed window lengths and gradient vanishing issues, which hindered their ability to model complex long-term dependencies \cite{B1}.

A paradigm shift emerged in 2017 with the introduction of the Transformer architecture proposed by Vaswani et al. \cite{B2}. The core innovation lies in its self-attention mechanism, which dynamically assigns weights to capture global dependencies across sequences. This approach circumvented the gradient vanishing problem inherent in LSTMs and GRUs while eliminating distance constraints in sequence modeling. The adaptability of Transformer spurred novel applications in time series analysis. For example, Lim et al. \cite{A8} proposed the Temporal Fusion Transformer integrating temporal feature encoding with interpretable attention heads, which achieved a 19\% reduction in mean absolute error compared to conventional LSTMs when dealing with retail demand forecasting. Similarly, Chen et al. \cite{B4} devised a graph-enhanced Transformer for multivariate time series anomaly detection, effectively modeling spatio-temporal dependencies and demonstrating superior performance over traditional methods in detection accuracy.

Despite these advances, technical progress during this period remained fragmented. These models were largely tailored to specific tasks, lacking a cohesive framework for broader applicability. Early Transformer-based approaches in time series analysis also faced challenges such as limited model capacity, poor cross domain generalization ability, and high transmission costs.

\subsection{The LLM-driven Paradigm}

The rapid advancement of LLMs, exemplified by GPT-3 and PaLM, has significantly advanced time series analysis by leveraging their remarkable generalization capabilities. These models excel in few-shot and zero-shot learning scenarios, achieving strong performance in different downstream tasks \cite{B6}. By pre-training on massive datasets, LLMs acquire the ability to infer patterns and reason with minimal task-specific examples, thereby reducing reliance on extensive labeled data. Furthermore, the emergence of cross-modal learning frameworks has further expanded the applicability of LLMs to time series analysis.

Recent research has explored adapting LLMs to numerical time series analysis. Gruver et al. \cite{A6} introduced an innovative approach to encoding time series into numerical token sequences, effectively reframing time series forecasting as a next-token prediction task. Experiments have shown that compared to specialized models trained specifically for time series, LLMs such as GPT-3 and LLaMA-2 achieve competitive or even higher performance in zero-shot time series extrapolation. Similarly, Sun et al. \cite{B8} proposed the Time Series Embedding via Symbolic Tokenization framework (TEST), which bridged the modality gap between time series data and LLMs by discretizing time series into symbolic sequences. By aligning these sequences with LLM-compatible embeddings, TEST enables pre-trained LLMs to interpret and process temporal patterns effectively, achieving state-of-the-art results in multiple benchmarks.

Building on these foundations, lightweight fine-tuning techniques have emerged as a prominent research to improve the adaptability of LLMs to time series tasks. For instance, Jin et al. \cite{B9} developed the Time-LLM framework, which integrated Low-Rank Adaptation and only fine-tuned 0.1\% of the model parameters. The approach finally achieved an accuracy of 92.1\% in the MIT-BIH arrhythmia classification task, emphasizing the potential of minimal but targeted adjustments to unlock domain-specific functionalities in general-purpose LLMs.

In addition, the rise of multimodal learning represents another key advancement in this field, as people are increasingly striving to combine time series with complementary modalities such as images, audio, and text for joint modeling. For example, Zhang et al. \cite{B10} introduced Insight Miner, a large multimodal model designed to generate time series descriptions enriched with domain-specific knowledge. By fine-tuning on a fusion dataset of time series and textual information, Insight Miner outperforms state-of-the-art multimodal models in producing coherent and contextually grounded insights for time series analysis. However, critical challenges still exist, for example, models like Insight Miner exhibit numerical sensitivity, where minor perturbations of input data can lead to inconsistent outputs and inference latency, limiting real-time deployment in time-sensitive scenarios. Addressing these limitations will be essential to ensure LLMs' robustness and scalability.

\subsection{Native Temporal Foundation Models Stage}

So far, time series analysis is evolving towards the stage of native temporal foundation models (TFMs), whose core is to integrate the basic features of time series into architecture design and training paradigms, thereby breaking through the limitations of traditional methods. Researchers from both industry and academia have conducted a series of explorations around TFMs. For instance, Google's TimesFM \cite{B12} unifies eight time series tasks (e.g., forecasting, classification) through a modular encoder-decoder design; the Lag-Llama \cite{B13}, a decoder-only Transformer, achieves state-of-the-art probabilistic forecasting via pretraining on cross-domain data and lag-based covariates. The research of UniTS \cite{B15} integrates multiple timing tasks into a single framework through task tokenization techniques with timing feature encoding module, outperforming 66 dedicated baseline models on 38 datasets, and marking a paradigm shift in timing model design from ``generic adaptation'' to ``native customization''.

Researchers have developed several innovative solutions to improve the LLMs' ability in dealing with long-sequence modeling. For example, the ProSparse Attention \cite{A7} enhances long-sequence modeling efficiency with attention pattern sparsification; the Performer \cite{B19} uses pyramidal attention structures to balance the short and long-term dependency capture with linear computational complexity. 

Additionally, few-shot learning is another crucial direction for enhancing TFMs' cross-domain generalization capabilities. Whereas conventional deep learning requires extensive labeled datasets, TFMs circumvent data limitations through meta-learning, self-supervised pretraining, and physics-informed fine-tuning. For example, the gradient-based meta-learning framework \cite{B61} achieves 18\% higher classification accuracy among 41 time series datasets, which establishes new benchmarks for time series classification. The Time-FFM \cite{B60} pioneers data privacy preservation through temporal-to-textual modality conversion and dynamic prompt generation, achieving a 85\% accuracy in zero-sample weather prediction.

Despite promising progress, TFMs also face challenges from theory to implementation. Firstly, the trade-off between efficiency and accuracy in long-sequence modeling remains unresolved, as sparse strategies may overlook critical temporal patterns. Secondly, integrating domain-specific priors such as physical laws and medical constraints, into general-purpose LLMs has not yet been developed, occasionally resulting in unrealistic predictions. Thirdly, real-time inference in high-frequency scenarios such as financial transactions requires further improvement in model compression. Future efforts may prioritize multimodal alignment and interpretability enhancement to bridge the gap between research prototypes and industrial grade systems, unlocking powerful and scalable solutions of LLMs for time series analysis.

\section{The Technical Landscape of LLM-driven Time Series Analysis}

\begin{figure}
\centering
\includegraphics[width=9 cm]{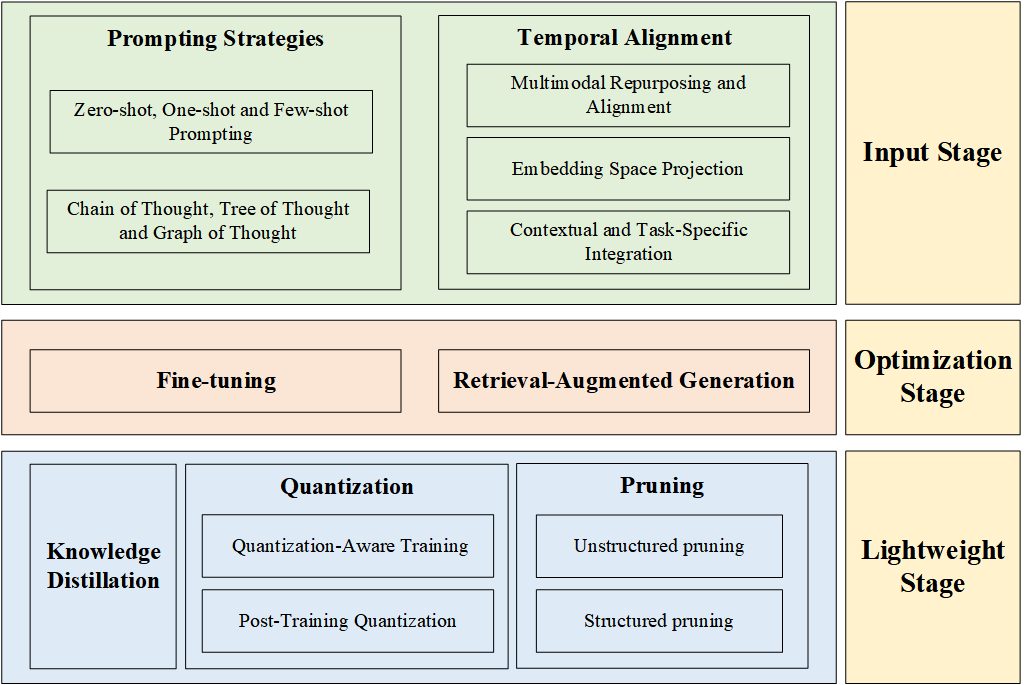}
\caption{The technical taxonomy of LLM-driven time series analysis.}
\label{fig3.1}
\end{figure}

To fully exploit the generalizability of LLM-driven time series analysis, this section provides an overview of LLM-driven time series analysis techniques and organizes a technical landscape according to the workflow perspective, namely LLMs' input, optimization, and quantification stages, as shown in Figure \ref{fig3.1}.

At the LLMs' input stage, techniques mainly enable processing time series data, in which prompting strategies and temporal alignment serve as significant roles. At the optimization stage, fine-tuning helps to achieve LLMs' parameter adaptation to specific time series patterns (e.g., financial fluctuations, sensor signals), and the Retrieval-Augmented Generation (RAG) mechanism can be introduced to dynamically infuse domain-relevant temporal patterns from external knowledge bases, so as to enhance LLMs' capability in time series analysis. At the lightweight stage, knowledge distillation, quantization and pruning can help alleviate the complexity and deployment costs of LLMs, especially in cases of edge computing and real-time scenarios.

\subsection{Techniques at LLMs' Input Stage}  

At LLMs' input stage, two techniques are employed, namely promoting strategies and temporal alignment methods. Promoting strategies enhance LLMs' ability to analyze time series by integrating time-aware semantic context through carefully designed prompt templates. These strategies address challenges in task adaptability, guiding LLMs to prioritize causal relationships and temporal patterns inherent in time series analysis. Temporal alignment, applied during data preprocessing, transforms raw time series data into structured formats compatible with LLMs. This technique resolves format compatibility issues, ensuring LLMs can effectively process mathematical properties and semantic features embedded within the time series.

\subsubsection{Prompting strategies}

Prompting strategies design clear and targeted instructions to communicate effectively with LLMs and enhance their capabilities on diverse time-series tasks. By crafting specific and detailed instructions, or ``prompts'', pre-trained LLMs can adapt to specific challenges and deliver accurate results without requiring extensive task-specific training data. The more detailed and structured the prompts are, the better LLMs can understand and fulfill the requests.

\paragraph{Zero-shot, One-shot and Few-shot Prompting}

Among all prompting strategies, zero-shot prompting \cite{C1, B6} serves as the simplest one, where only task descriptions need to be provided with LLMs and task reasoning can be performed directly. Compared with zero-shot prompting, one-shot and few-shot prompting are techniques based on In-Context Learning (ICL), which allow LLMs to learn from examples embedded directly in the prompts instead of additional training or fine-tuning. They could enhance zero-shot prompting by providing a single or few examples before the new task, and help clarify expectations and improve LLMs' performance.

With the iterative upgrade of LLMs, they have the ability to handle special tasks by zero-shot, one-shot and few-shot prompting. For instance, Brown et al. \cite{B6} found that GPT-3 with 175 billion parameters demonstrated powerful few-shot learning capabilities without gradient updates or fine-tuning. Models such as BLOOM \cite{C3-} and LLaMA \cite{C4,C5} typically fine-tuned for different target tasks, also exhibit strong few-shot and zero-shot learning capabilities without fine-tuning.

At present, some scholars have explored the application of zero-shot, one-shot and few-shot prompting in LLM-driven time series analysis. For example, Xue et al. \cite{C6} proposed PromptCast, a prompt-based method specifically designed for time series forecasting tasks. The experiments show that PromptCast exhibits better generalization ability in zero-shot scenarios and significantly outperforms traditional numerical prediction methods in urban temperature, power load, and human mobility prediction tasks. Jin et al. \cite{B9} demonstrated that LLMTime could stimulate the zero-sample time series learning ability by providing additional contextual information and task descriptions. Hu et al. \cite{C8} combined EEG data with facial expressions and experimentally showed that one-shot prompting can better tap the potential of LLMs in mental health assessment than zero-shot learning. Gruver et al. \cite{A6} also found that zero-shot prompting of time series prediction by LLMs could outperform dedicated time series models trained for downstream tasks.

The zero-shot, one-shot and few-shot prompting strategies can achieve rapid task adaptation by flexibly calling the prior knowledge of pre-trained LLMs, without data annotation and model fine-tuning. However, such methods are limited by pre-trained LLMs' knowledge timeliness, the sensitivity of prompts, and the shortcomings of complex reasoning, which are difficult to use in professional, dynamic, and complex scenarios. It still needs to be combined with hybrid architectures such as retrieval enhancement, fine-tuning technology, or chain of thought to break through bottlenecks \cite{C46,C47}.

\paragraph{Chain of Thought, Tree of Thought and Graph of Thought}

Chain of Thought (CoT) is another popular prompting strategy that simulates human-like reasoning through a series of intermediate steps. It aims to break down problems into manageable thoughts that sequentially lead to a conclusive answer. Compared with the ICL mentioned above, CoT not only provides prompt examples but also provides the thinking steps and paths. In this case, it can better guide LLMs to think and reason properly. This prompting strategy has shown great potential in LLMs' tasks. For example, Wei et al. \cite{C10} demonstrated that a few CoT examples can enhance the performance of LLMs on arithmetic and commonsense reasoning tasks. Zhang et al. \cite{C11} integrated visual features with LLMs to propose Multimodal-CoT, employing a decoupled training framework that generates effective reasoning chains during inference. The results show that it exceeds GPT-3.5 on the ScienceQA benchmark with less than 1 billion parameters.

Shortly after CoT, the Tree of Thought (ToT) \cite{C12} was proposed which transforms the reasoning task into a tree search problem, and the intermediate tree states are partial solutions to which LLMs refer. At the same time, the prompting strategy of Graph of Thought (GoT) \cite{C13} was also generalized as an arbitrary graph structure, rather than just depending on the tree structure. These prompting strategies have been adopted in time series analysis tasks. For example, Kim et al. \cite{B37} applied ToT to clinical epilepsy EEG analysis and showed that the strategy was effective in LLM-driven EEG abnormality detection. Ho et al. \cite{C15} adopted CoT for fine-tuning student models with reasoning samples generated by teacher models, and experiments demonstrated that the proposed method significantly improves the ability of smaller student models in complex arithmetic and commonsense reasoning tasks. 

The CoT, ToT and GoT are progressive prompting strategies that help LLMs solve complex problems. The CoT decomposes problems through easily traceable linear steps, while it is limited to a single path with low fault tolerance and high difficulties to backtrack. The ToT introduces bifurcation exploration, allowing for parallel derivation of multiple paths and increasing the possibility of global optimal solutions, but with high computational costs and complex path management. The GoT is further expanded into a mesh structure, supporting flexible associations and circular reasoning between nodes. It excels at handling nonlinear problems, but is difficult to implement due to high resource costs and poor interpretability. Each prompt strategy has its advantages and disadvantages, and it is necessary to consider practical application requirements, and choose the appropriate one by achieving the balance of efficiency, flexibility, and complexity.

\subsubsection{Temporal Alignment}

LLMs for time series analysis suffer a lot from temporal misalignment issues, therefore, it is necessary to transform time series into representations suitable for LLMs \cite{C48}. 

Most research focuses on multimodal repurposing to activate LLMs' capabilities by aligning time series with the modalities of pre-trained tasks \cite{B46}. For example, Wang et al. \cite{C19} attempted to align EEG signals with texts, and combined with a pre-trained BART model to decode and generate text. Jin et al. \cite{B9} designed a reprogramming layer to align time series with text embeddings based on a pre-trained LLM. The work enhanced the LLM's ability of time series analysis by introducing contextual information and task instructions. Inspired by this work, Chan et al. \cite{B32} improved it and proposed the MedTsLLM framework by combining multivariate time series signals with texts, and provided rich contextual information for LLMs. Experiments show that MedTsLLM can be effectively applied to semantic segmentation, boundary detection and anomaly detection tasks of physiological signals.

Although existing studies have addressed the temporal alignment problem for LLM-driven time series analysis through a variety of techniques and architectural adjustments, a deeper integration of temporal patterns with the transformer architecture of LLMs remains an open challenge, suggesting that future directions should focus on developing more sophisticated alignment paradigms to better coordinate time series data with the core architectural advantages of LLMs.

\subsection{Techniques at LLMs' Optimization Stage}

At LLMs' optimization stage, two techniques are concluded to improve LLMs' ability of time series analysis, namely fine-tuning and the Retrieval-Augmented Generation (RAG). The fine-tuning refers to adapting LLMs' parameter according to specific tasks, enabling LLMs to capture time-dependent features. The RAG techniques dynamically integrate external temporal knowledge (e.g., domain-specific databases, historical patterns) during inference, enriching context-aware predictions by retrieving and incorporating relevant temporal dependencies.

\subsubsection{Fine-tuning}

Fine-tuning is the process of adjusting a pre-trained LLM's parameters with smaller and task-specific datasets. By fine-tuning, the LLMs can be optimized for specific tasks while retaining the pre-trained knowledge, thus improving the task-specific performance. According to the purpose of fine-tuning, existing LLM fine-tuning methods for time series analysis can be divided into fine-tuning specialized-domain models and fine-tuning general-purpose models.

For specialized-domain models, it means the LLMs will be fine-tuned with time series data from specific domains (e.g., finance, healthcare). For example, Yu et al. \cite{B28} fine-tuned GPT-4 by combining time series data of stock prices, to realize the prediction of future stock ups and downs. Wang et al. \cite{Cadd2} fused news events and time series data, and fine-tuned a pre-trained LLM to learn the correspondence between time series and news events more efficiently. Kim et al. \cite{Cadd3} combined a context-enhancement strategy with time-dependent physiological health data and fine-tuned the LLM for health prediction tasks, which outperformed larger-scale models a lot on multiple tasks.

While specialized-domain models perform well in specific tasks, the demand for general-purpose models is increasingly evident when faced with diverse time series tasks. Fine-tuning general-purpose models \cite{B48,Cadd5,Cadd6,Cadd7} is capable of achieving higher performance in various tasks, providing a more versatile solution for time series analysis. For example, Zhou et al. \cite{B25} employed pre-trained models based on massive text and image data, and adapted to time series tasks through fine-tuning. The experiments showed that the method achieved significant improvements in time series classification and anomaly detection. Cao et al. \cite{B34} decomposed time series into trend, seasonality, and residual components, and then encoded each component separately to guide LLMs better understanding the complex structure of time series. Aghakhani et al. \cite{Cadd8} fine-tuned LLaMA 2 for time series forecasting, which solved the problem of data scarcity and poor model generalization ability. Liu et al. \cite{Cadd9} established the CALF framework by introducing cross-modal matching modules, feature regularization loss, and output consistency loss, significantly improving the prediction accuracy in multiple long-term and short-term prediction tasks.

Overall, fine-tuning has played an important role in LLM-driven time series analysis. On the one hand, fine-tuning can fully utilize the language understanding and representation capabilities of pre-trained LLMs, and achieve efficient performance improvements in tasks such as finance and medicine by introducing domain specific data. On the other hand, it can alleviate the challenges brought by complex tasks and cross-domain requirements, and reduce the cost of training models repeatedly for different tasks. To be noted, fine-tuning also has certain limitations. For example, specialized-domain models often lack flexibility and are difficult to adapt to new time series tasks, while general-purpose models are easily affected by significant structural differences in time series data, leading to performance fluctuations \cite{A1}. In addition, it is still necessary to approach the requirements in terms of computing resources and labeled data when fine-tuning LLMs. In the future, lightweight fine-tuning strategies and efficient data utilization methods can be further explored to improve the processing capability of fine-tuning in LLM-driven time series analysis.

\subsubsection{Retrieval-Augmented Generation}

Even with fine-tuning, LLMs still face challenges when dealing with time series data. On the one hand, static knowledge is difficult to capture the dynamic evolution trend of time series. On the other hand, the real-time nature of training data leads to LLMs' knowledge lag, especially in scenarios such as financial forecasting and health monitoring. 

The RAG technology \cite{C27} was originally proposed by Meta AI and has been proven to be highly effective in improving the credibility and diversity of language generation tasks. So far, it has been increasingly applied to LLM-driven time series analysis. By real-time retrieval of external time series databases, time series can be dynamically injected into the generation process in a timely manner, enabling LLM to reason based on the latest context. RAG not only effectively alleviates the problem of knowledge lag, but also avoids the risk of sensitive data leakage through localized retrieval with permission control.

For example, Tire et al. \cite{C28} introduced a principled RAG framework for time series forecasting, called Retrieval Augmented Forecasting (RAF), and developed efficient strategies on this framework to retrieve relevant time series examples, resulting in significant improvements in prediction accuracy across different time series domains. Zhang et al. \cite{C29} designed an end-to-end trainable retriever that can extract valuable information from a customized time series knowledge base, achieving better performance in multiple fields such as energy, transportation, and health. Yang et al. \cite{B27} constructed a time series knowledge base from historical sequences and used dynamic time regularization to retrieve reference sequences with similar patterns to the query sequence, which served as input of LLMs and effectively improved the prediction accuracy of time series.

RAG can to some extent solve the knowledge lag and dynamic adaptation problems of LLM-driven time series analysis. However, RAG still faces some limitations in time series applications. On the one hand, the performance of the retrieval module largely depends on the quality and update frequency of the knowledge base \cite{Cadd11}. On the other hand, the integration of external knowledge bases in terms of format conversion, semantic understanding, and reasoning ability also faces challenges \cite{Cadd12}. In the future, research should further explore constructing structured time series knowledge bases with comprehensive semantic annotations, to enhance the retrieval precision and data interpretability \cite{Cadd13}.

\subsection{Techniques at LLMs' Lightweight Stage}

LLMs has achieved significant performance improvements in various tasks. However, the large pre-trained models represented by the Transformer architecture usually contain hundreds of millions or even billions of parameters, with high computational complexity and extremely high requirements for hardware resources. In practical applications, devices such as mobile terminals, edge nodes, etc. are strictly limited in computing power, storage capacity, and energy budget. Therefore, lightweight deployment of LLMs has become one of the hot topics \cite{C31}. This section systematically explores three popular lightweight techniques for LLM-driven time series analysis, namely the knowledge distillation, quantization, and pruning.

\subsubsection{Knowledge Distillation}

Knowledge distillation is a technique that transfers the knowledge of pre-trained teacher models to smaller student models. It adopts teacher models' output distribution or intermediate features as supervisory signals to guide the training of student models, thereby reducing the models' size while maintaining high performance.

Depending on the degree of access to internal information of teacher models, knowledge distillation can be divided into black-box distillation and white-box distillation. The black-box distillation means that student models only know the prediction results of teacher models for specific input data, instead of the internal structure, parameters or intermediate layer output, while the white-box distillation means that student models can access the internal structure, parameters and intermediate layer information of teacher models. So far, knowledge distillation has been widely used in the field of NLP \cite{C15,C32,C33}, and has begun to emerge in large-scale time series models.

For example, Xu et al. \cite{C34} proposed a contrastive adversarial knowledge distillation framework that uses adversarial adaptive automatic alignment to synchronize feature distributions when student and teacher models have different architectures, while combining instance-level feature alignment with contrastive loss to address the limitations of traditional knowledge distillation in time series regression tasks. Campos et al. \cite{C35} proposed a LightTS framework based on adaptive ensemble distillation technology, which can compress large ensembles into lightweight models while ensuring competitive precision. Ni et al. \cite{C36} presented a cross framework knowledge extraction method that combines lightweight multilayer perceptrons with advanced Transformer and CNN architectures, which not only reduces computational and storage costs, but also achieves better performance in long-term time series prediction.

Knowledge distillation has shown growing potential in LLM-driven time series analysis by enabling efficient model compression and cross-architecture knowledge transfer. It helps reduce computational costs while retaining performance in order to support lightweight deployment. Future directions may focus on developing lightweight LLM-specific knowledge distillation strategies, integrating self-supervised pre-training to reduce teacher dependency, and designing temporal-aware alignment mechanisms. 

\subsubsection{Quantization}

Quantization is the process of reducing the bit width of LLMs' parameters and minimizing the inference loss. It is primarily categorized into Quantization-Aware Training (QAT) and Post-Training Quantization (PTQ). QAT simulates quantization operations during model training, allowing it to maintain a high accuracy rate, but the training process is more complicated; PTQ, on the other hand, performs quantization after model training is completed, which is suitable for scenarios that require rapid deployment.

In time series analysis, quantization has begun to be adopted for model lightweight. For example, Kermani et al. \cite{C42} systematically explored the impact of quantization on Transformer-based time series classification models, and the results showed that static quantization reduced energy consumption by 29.14\% while maintaining classification performance. Liu et al.\cite{C37} proposed a quantization-aware training method combined with data-free distillation, allowing the quantization model to better preserve the original output distribution.

Quantization is crucial in the deployment of LLM-driven time series analysis, especially in resource-constrained scenarios, as it can significantly reduce computing and memory overhead and improve the inference speed. The future direction may develop towards exploring a lightweight dynamic quantization framework based on hybrid quantization strategies to further optimize the practical deployment.

\subsubsection{Pruning}

Pruning is another popular method for model lightweight, which reduces the size and computational complexity of models by removing unimportant connections or neurons. Pruning techniques can be divided into unstructured pruning and structured pruning. Unstructured pruning removes connections or neurons at any position of the models, while structured pruning reduces models' complexity by removing entire neurons or convolutional layers. In the field of NLP, scholars have paid great attention to the pruning strategies of LLMs, such as unstructured pruning method for large-scale GPT models proposed by Frantar and Alistarh \cite{C43}, loss function-based structured pruning method proposed by Molchanov et al. \cite{C44}, and gradient-based one-time structured pruning method proposed by Ma et al. \cite{C45}.

Due to the increasing scale and complexity of time series, some researchers have also explored the application of LLM pruning techniques to reduce computational and storage costs when conducting time series analysis. 

For example, Kermani et al. \cite{C42} applied magnitude-based and layer-by-layer pruning techniques to Transformer models for time series classification tasks, and the experiments demonstrated that the pruning techniques can help improve inference speed by up to 63\%. Jin et al. \cite{Cadd15} only retained the ``knowledge'' related to the task by pruning, avoiding loading the entire LLM in the training and reasoning stages, which can reduce the consumption of computing resources, and is suitable for resource-constrained scenarios.

Knowledge distillation, quantization, and pruning are popular techniques at LLMs' lightweight stage. Table \ref{cab1} compares them from aspects of main tasks, calculation overhead and suitability. Knowledge distillation enables cross-architecture knowledge transfer and is effective for training small models with high performance, while it typically requires joint training and leads to higher computational costs \cite{Cadd16,Cadd17}. Quantization maps model weights and activations to low-bit representations, making it highly suitable for hardware deployment by significantly lowering the storage and energy consumption. Pruning reduces the size of models by removing redundant parameters or structures, offering good hardware adaptability while maintaining the original architecture. In general, the above methods have their own advantages and disadvantages, and it is necessary to determine an appropriate lightweight technique based on the practical requirements.

\begin{table*}[ht]
\centering
\caption{Comparison of Lightweight Strategies in LLMs-Driven Time Series Analysis}
\label{cab1}
\begin{tabular}{>{\centering\arraybackslash}m{4cm} 
                >{\centering\arraybackslash}m{4cm} 
                >{\centering\arraybackslash}m{4cm} 
                >{\centering\arraybackslash}m{4cm}}
\toprule
\textbf{Strategy} & \textbf{Core Objective} & \textbf{Calculation Overhead} & \textbf{Suitability} \\
\midrule
Knowledge Distillation & Cross-architecture compression and knowledge transfer & High & High \\
Quantization & Numerical compression of parameters and activations & Low (PTQ Requires No Retraining) & Low \\
Pruning & Model sparsification by removing redundant components & Medium & Medium \\
\bottomrule
\end{tabular}
\end{table*}

\section{Popular LLMs enhancing time series analysis}%

\begin{figure}
\centering
\includegraphics[width=9 cm]{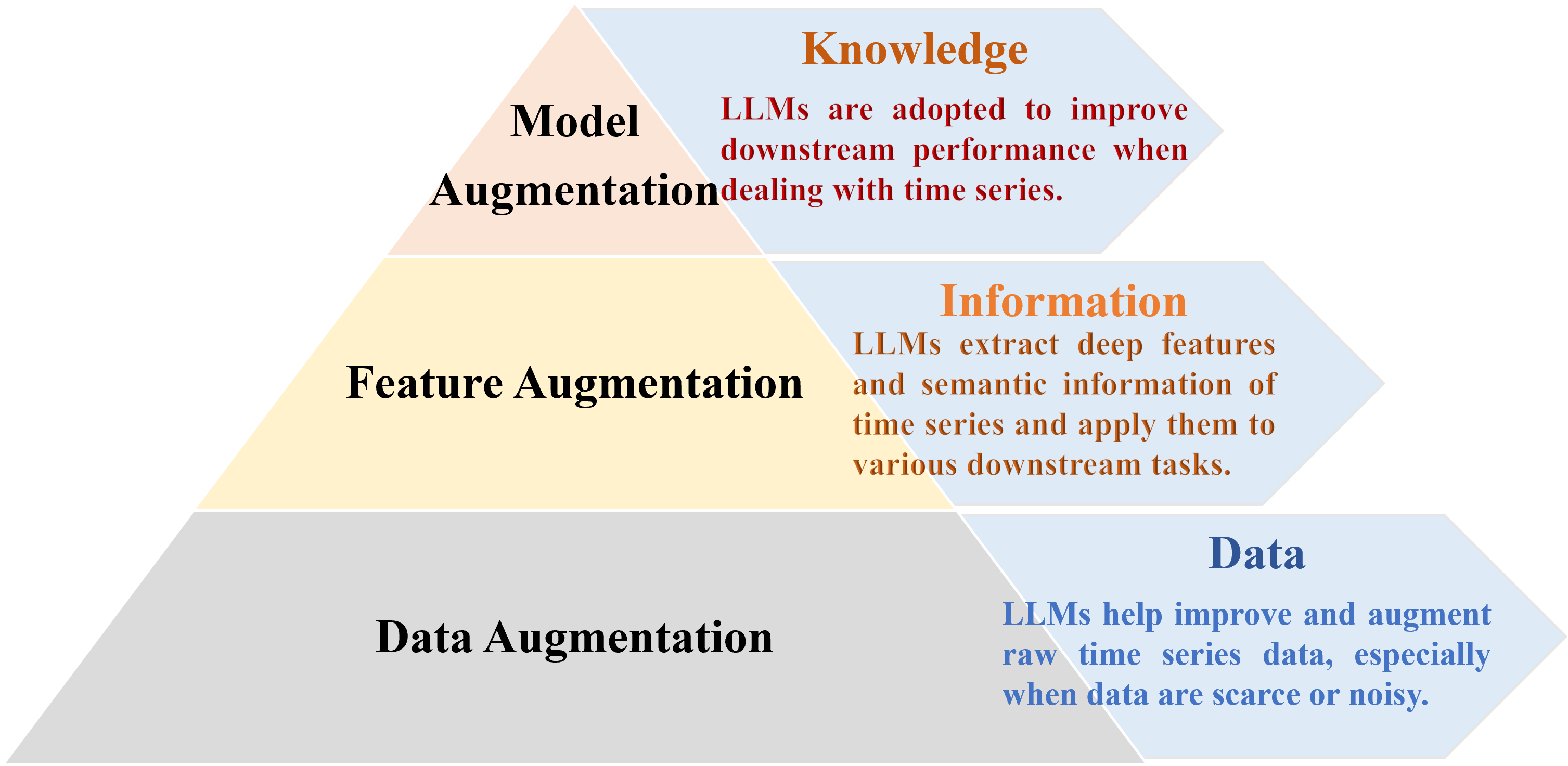}
\caption{The structured hierarchy describing LLMs' augmentation in time series analysis.}
\label{Fig1}
\end{figure}

Inspired by the theoretical framework Data-Information-Knowledge-Wisdom (DIKW) \cite{B23} in information science, we organize popular LLMs in time series analysis into three progressive levels: Data augmentation (Data), Feature augmentation (Information), and Model augmentation (Knowledge). As shown in \ref{Fig1}, the structured hierarchy demonstrates how LLMs connect granular signal processing with strategic decision making, thus achieving a coherent transition from raw time data to domain intelligence.

\subsection{Data augmentation}

At the data augmentation level, we focus on how LLMs help improve and augment raw time series data, especially when data are scarce or noisy. With the generative capability of LLMs, it is possible to generate new time series data that are similar to the original ones, or match and merge time series with other sources of information, thus expanding the diversity of the datasets and improving data quality. The types of data augmentation can be divided into generation-based and fusion-based data augmentation.

\subsubsection{Generation-based Data Augmentation}

Due to the powerful generative and contextual understanding capabilities of LLMs, they can help generate synthetic time series data that preserve structural and semantic similarities to the original ones, addressing challenges in scenarios where data are scarce or annotation costs are prohibitive.

Recent advances in leveraging LLMs for time series analysis demonstrate their growing versatility. Zhou et al. \cite{B24} developed a cross-domain collaborative framework using pre-trained LLMs to generate synthetic time series data, addressing data scarcity and enhancing diversity in intelligent systems. Xie et al. \cite{B25} introduced ChatTS, a novel method that aligns time series with textual semantics to generate structurally consistent synthetic data, improving analytical depth and prediction accuracy in data-limited scenarios. Similarly, Lee et al. \cite{B26} proposed LLM2LLM, an iterative framework that augments training datasets through multi-round synthetic data generation and feedback-driven refinement, boosting LLMs' learning capabilities in low-resource settings. Yang et al. \cite{B27} advanced this field with TimeRAG, a retrieval-augmented framework that builds a knowledge base on time series from historical data. By dynamically retrieving reference sequences and integrating them as contextual prompts into LLM-based predictions, TimeRAG significantly improves forecast accuracy. 

The integration of LLMs into time series analysis marks a paradigm shift in addressing data scarcity and enhancing analytical robustness. By leveraging the generative prowess, LLMs not only synthesize structurally and semantically coherent time series data but also enable cross-domain knowledge transfer. The generation-based data augmentation approaches alleviate annotation bottlenecks and improves model generalization in low-resource settings.

\subsubsection{Fusion-based Data Augmentation}

In LLM-driven time series analysis, the fusion-based data augmentation techniques break through limitations of traditional methods by deeply integrating time series with other modal data such as images, text, event logs, etc. \cite{B29}. The core is to use the cross-modal semantic understanding ability of LLMs and build a dynamic mapping relationship between time series and auxiliary information, generating enhanced datasets with expanded dimensions and rich semantics. 

In medical scenarios, physiological signals (ECG, EEG) can be jointly encoded with patient medical records and medical images, and the potential correlation between pathological characteristics and text descriptions can be captured through cross-modal attention mechanisms. For example, Chan et al. \cite{B32} proposed a multimodal medical time series analysis system based on MedTsLLM, which efficiently integrates time series data from various medical devices and converts these data into a unified representation format to improve the accuracy of analysis and prediction. 

Additionally, some research has been demonstrated to align and fuse multimodal datasets with time series, in order to improve models' ability. For instance, Tao et al. \cite{B30} established a hierarchical multimodal large model HiTime, which effectively combines dynamic temporal features with text semantics, thereby improving the model's processing and analysis capabilities. Pan et al. \cite{B31} proposed LLM-based semantic space notification cue learning, which aligns time series embedding with the pre-trained semantic space, and performs time series prediction based on the cues learned from the joint space. Ge et al. \cite{B35} established the WorldGPT, which extends LLM into a multimodal model that can simultaneously integrate information from different sources (such as images, videos, audio, sensors, etc.), thereby enhancing the model's understanding and reasoning capabilities.

The LLM-driven fusion-based data augmentation has opened up an innovative path for time series analysis. It not only breaks through the data limitations of a single modality, but also mines potential semantic relationships among various modalities, significantly improving the generalization ability of models in scenarios where samples are scarce.

\subsection{Feature augmentation}

This section focuses on the contribution of LLMs in feature augmentation, especially how to use LLMs to extract deep features and semantic information of time series and apply them to various downstream tasks. Depending on the application scenarios, the techniques can be divided into semantic understanding, context awareness, and sequence modeling.

\subsubsection{Semantic understanding}

In time series analysis, traditional feature extraction methods usually focus on numerical and statistical features while ignoring deep semantic structures, while the LLM-driven semantic understanding can help understand deeper semantics and augment more advanced features.

So far, semantic understanding has become the main research direction in LLM-driven time series analysis. For example, Zhang et al. \cite{B36} adopted popular LLMs like GPT-3.5 Turbo, GPT-4, and LLaMA to extract complicated features from EEG signals and eye-tracking data. It was demonstrated that the identification accuracy of neural states could be improved largely during reading comprehension tasks. Similarly, EEG-GPT \cite{B37} provides an innovative approach to converting EEG signals into the form of ``language sequences'' and explores more semantic patterns within the signals. It can capture the complex relationship between temporal patterns and semantic information in EEG data for in-depth semantic understanding, thus improving the classification and inference of EEG activity.

Since LLMs are pre-trained with massive data and have deep context modeling capabilities, they show significant advantages in semantic understanding when dealing with time series data. Specifically, LLMs excel at identifying intricate patterns and long-term dependencies in time series data, moving beyond superficial numerical analysis to uncover the underlying significance of temporal fluctuations. It has been proven immeasurable value in fields such as financial market forecasting and weather forecasting, which require a deep understanding of time dynamics. By harnessing nuanced insights embedded in time series, LLMs generate highly accurate and interpretable predictions, empowering data-driven decision-making with actionable intelligence.

\subsubsection{Context awareness}

In traditional feature extraction methods, features such as mean, variance, and other static measures are usually fixed and do not fully utilize the temporal relationships and contextual information present in the data. In contrast, LLMs, especially those based on the Transformer architecture (e.g. BERT), can dynamically adapt the embedding representation to the context to produce richer features. This context-aware embedding not only processes data at a single point in time, but also considers the temporal context of the preceding and following steps, capturing complex dependencies in extended time series and further enabling a more robust representation of the inherent structure in time series data.

Recent research has explored LLM-driven context-aware methods for time series analysis. For instance, TS2Vec \cite{B38} introduces a contrastive learning framework designed to learn multi-scale contextual representations of arbitrary subsequences within time series. By hierarchically augmenting contextual views, the model captures nuanced semantic relationships across varying temporal granularities, and demonstrates versatility across diverse applications such as forecasting, classification, and anomaly detection. Shao et al. \cite{B39} presents a novel framework, named STEP, which integrates a pre-training model with Spatial-Temporal Graph Neural Networks, to address the challenge of modeling long-term dependencies in multivariate time series forecasting. It highlights that context-aware modeling, akin to language models' ability to process sequential semantics, is equally vital for time series where long-term context drives accurate predictions. 

The context-aware methods powered by LLMs offer a significant advancement in time series analysis by moving beyond traditional fixed-feature extraction toward dynamic, context-driven representations. It enables to effectively capture temporal dependencies and contextual information across multiple scales, supporting more nuanced understanding of time series. Context awareness is becoming a pivotal paradigm in time series analysis, bridging the gap between time series data and semantic understanding.

\subsubsection{Sequence modeling}

Sequence modeling is another key advantage of LLM-driven time series analysis \cite{B40}. Unlike conventional statistical methods struggling to capture long-term dependencies inherent in time series data, LLMs excel at flexibly modeling multiscale temporal dynamics. By analyzing interrelationships across entire sequences, they simultaneously identify both short-term fluctuations and long-range patterns, and enable to derive robust insights from time series.

Many studies have explored techniques such as multimodal fine-tuning \cite{B41} and model reprogramming \cite{B42} and demonstrated how LLMs can be structurally adapted for sequence modeling. A notable example is the Time-LLM framework proposed by Jin et al. \cite{B9}, which reformulates time series into text-based prototypes aligned with the strengths of LLMs. By integrating the self-attention mechanisms inherent in LLMs, this approach enables LLMs to effectively identify critical temporal patterns while preserving contextual coherence. Similarly, Yang et al. \cite{B44} introduced Voice2Series which reimagines time series as acoustic signals, enabling cross-domain transfer of speech recognition techniques to time series classification. By aligning time series with acoustic modeling frameworks, the method leverages LLMs' inherent strengths in hierarchical temporal pattern recognition. Chen et al. \cite{B45} developed the fLLM-TS framework, which integrates time series embeddings with LLM-generated semantic representations through mutual information maximization. It not only enhances feature extraction but also enables LLMs to infer latent dependencies across extended sequences, advancing applications like industrial predictive maintenance and multi-horizon forecasting.

Sequence modeling represents a powerful capability of LLMs in time series analysis, enabling the capture of both short-term fluctuations and long-range dependencies that traditional methods often overlook. These advancements highlight the transformative potential of LLMs in extracting meaningful insights from complex sequential data, paving the way for more accurate time series analysis across diverse domains.

\subsection{Model augmentation}

At the model augmentation level, we analyze how LLMs are adopted and optimized to improve downstream performance when dealing with time series. Traditional time series models often rely on fixed architectures that struggle to adapt to the complexity and variability inherent in real-world time series data \cite{B46}. To overcome these limitations, researchers have turned to LLMs for model augmentation. By integrating the semantic understanding and contextual reasoning capabilities of LLMs with conventional time series frameworks, hybrid architectures can capture both localized temporal patterns and broader contextual influences, thereby improving the predictive accuracy and interpretability \cite{B62}. 

A key innovation in this domain is the dual stream architecture that operates traditional models and LLMs in parallel to process distinct aspects of time series \cite{B63, B64}. On the one hand, traditional models such as LSTM \cite{B65} or ARIMA \cite{B66} focus on extracting structural patterns such as short-term trends, seasonality, and autoregressive dependencies, and they excel at processing numerical sequences and identifying temporal relationships within the data. On the other hand, LLMs \cite{B67} can help analyze auxiliary textual or contextual information such as news articles, financial reports, or event logs, to uncover latent semantic features, sentiment shifts, or external events that might influence the time series. The outputs from both streams are then fused, creating a unified representation that combines numerical precision with contextual awareness. It ensures that both the rigid temporal dynamics and the fluid external factors could be accounted for, enhancing robustness in scenarios where data variability or incomplete modalities might otherwise degrade performance.

To harmonize predictions from these dual components, ensemble learning strategies are often employed. Boosting methods \cite{B68}, for instance, iteratively refine predictions by combining weak learners (e.g., ARIMA) with LLM outputs, prioritizing corrections for mispredicted instances. Bagging techniques \cite{B69}, alternatively, leverage bootstrapped sampling to generate diverse model ensembles, reducing variance and improving stability. Stacking \cite{B70} takes a meta-learning approach, where a higher-level model learns to optimally blend predictions from base models (e.g., LSTM) and LLMs, often achieving superior generalization by exploiting complementary strengths. These strategies ensure that the integration of LLMs is not merely additive but synergistic, enabling the combined system to outperform individual components.

Recent studies highlight the versatility of LLM-augmented frameworks in diverse domains. One notable example is the work of Zhang et al. \cite{B56}, which integrates LLM with graph convolutional networks (GCN) through a multi-instructor knowledge distillation framework. This approach addresses spatio-temporal forecasting challenges by distilling semantic insights from LLMs into the GCN architecture. Another approach, exemplified by the ViLBERT \cite{B57} proposed by Lu et al., extends the BERT architecture to a dual-stream multimodal model capable of processing visual and textual inputs through cross-modal attention. It has inspired adaptations for time series analysis enriched with multimedia context, such as event-driven traffic or weather predictions. In financial analytics, the Text2TimeSeries framework \cite{B58} leverages LLMs to extract market sentiment and event-driven insights from textual data, which are fused with traditional forecasts to refine predictions. Similarly, in transportation prediction, Ma et al. \cite{B59} combine a pre-trained LLM with spatio-temporal transformers to improve traffic flow forecasting, particularly under the influence of external events like accidents or road closures.

These advancements are systematically categorized in Table \ref{tab1}, which classifies LLM-augmented models in aspects of scope, task, and architecture. 
The scope of the models is labeled as the general-purpose one or the specialized-domain one, depending on whether it addresses general time series challenges or specific applications in finance, healthcare, or transportation. 

Collectively, these efforts illustrate LLMs are transcending their traditional roles in NLP to become pivotal tools in time series analysis. By bridging numerical modeling with semantic reasoning, they enable systems to not only predict future trends but also interpret the underlying drivers of those trends. This dual capability positions LLM-augmented models as transformative paradigms for time series analysis, particularly in an era where data complexity and multimodal inputs demand both computational rigor and contextual agility.

\begin{table*}[!ht]
\centering
\footnotesize\caption{Models for time series analysis augmented by LLMs}
\label{tab1}
\begin{tabularx}{\textwidth}{>{\centering\arraybackslash}m{4cm} 
                >{\centering\arraybackslash}m{4cm} 
                >{\centering\arraybackslash}m{4cm} 
                >{\centering\arraybackslash}m{4cm}
                >{\centering\arraybackslash}m{4cm}
  }
    \toprule
    \makebox[0.05\linewidth][c]{\textbf{Models}} & \textbf{Scope} & \makebox[0.05\linewidth][c]{\textbf{Tasks}} & \textbf{Architecture} & \textbf{Year} \\
    \midrule
    Time-LLM\cite{B9} & General-purpose & Forecast & -- & 2023   \\ 
    LLM4TS\cite{B48} & General-purpose & Forecast & --  & 2023 \\ 
    TEMPO\cite{B34} & General-purpose & Forecast & --  & 2023 \\
    METS\cite{C24} & Specialized-domain & Healthcare & --  & 2023 \\
    LAMP\cite{B49} & Specialized-domain & Event prediction & --  & 2023 \\
    SigLLM\cite{B50} & General-purpose & Detection of Anomalies & --  & 2024 \\
    AnomalyLLM\cite{B51} & General-purpose & Detection of Anomalies & -- & 2024 \\
    Voice2Series\cite{B44} & General-purpose & Categorisation & --& 2021 \\ 
    MTSMAE\cite{B52} & General-purpose & Forecast & -- & 2022 \\
    TSMixer\cite{B53} & General-purpose & Forecast & -- & 2023 \\
    PatchTST\cite{B54} & General-purpose & Forecast & -- & 2023 \\
    PromptTPP\cite{B55} & Specialized-domain & Event prediction & -- & 2023 \\
    LM-KD\cite{B56} & Specialized-domain & Healthcare & 
    \begin{tabular}[c]{@{}c@{}}LLM-Graph Convolutional \\ Neural Network\end{tabular} & 2024 \\
    ViLBERT\cite{B57} & General-purpose & Categorisation & 
    \begin{tabular}[c]{@{}c@{}}Expansion of \\ the BERT model\end{tabular} & 2019 \\
    Text2TimeSeries\cite{B58} & Specialized-domain & Finances & 
    \begin{tabular}[c]{@{}c@{}}LLM - Modeling \\ of State Changes\end{tabular} & 2024 \\
    STTLM\cite{B59} & Specialized-domain & Predicting Flow of Traffic & PFM-STTN & 2024 \\
    \bottomrule
\end{tabularx}
\end{table*}

\section{Potential Applications}

\subsection{Anomaly Detection}

Anomaly detection, a critical application of time series analysis, focuses on identifying abnormal patterns to alert system failures, network attacks, or significant data shifts. The LLM-driven time series analysis relies on LLMs' powerful sequence modeling capabilities to learn rich feature representations, so as to realize anomaly detection accurately.

So far, studies are focusing on improving anomaly detection performance based on LLM-driven time series analysis. For example, the Transformer architecture proposed by Vaswani et al. \cite{B2} effectively captured long-range dependencies in time series, opening new perspectives for time series analysis. The T5 model proposed by Raffel et al. \cite{D2} further demonstrated the potential of LLM-driven time series analysis in unsupervised multi-task learning, laying the foundation for anomaly detection. The LLM-driven time series analysis could improve the anomaly detection performance due to the strong ability in feature extraction, contextual understanding and interpretability.

Despite their potential, the LLM-driven anomaly detection faces notable challenges. First, the substantial computational resources and memory required for LLMs' training and deploying pose practical barriers to real-world implementation. Second, the probabilistic nature of LLMs can lead to false positives or negatives, particularly when encountering rare or novel anomaly types. Third, integrating domain-specific knowledge into LLMs remains an unresolved issue, as current models often lack mechanisms to incorporate expert insights or contextual constraints effectively.

Future research should prioritize optimizing LLMs' architectures and algorithms to address these limitations. For example, the sparse attention mechanism \cite{D4} could reduce computational complexity while maintaining performance. To mitigate false predictions, hybrid frameworks \cite{D5} may improve robustness by leveraging complementary strengths across different approaches. Furthermore, techniques such as reinforcement learning or transfer learning \cite{D7} could refine parameter tuning and adaptability for various anomaly detection tasks, allowing LLMs to dynamically adapt to evolving data environments.

Although challenges such as computational cost, reliability, and domain adaptability still exist, the LLM-driven time series analysis represents a paradigm shift in the field of anomaly detection, providing advanced tools for processing complex temporal patterns. In future research, researchers can fully unleash the potential of LLMs to build accurate, scalable, and context-aware anomaly detection systems.

\subsection{Boundary Detection}

Boundary detection, a fundamental task in time series analysis, focuses on identifying the precise start and end points of events or pattern transitions within sequential data. It has gained significant advancements with the integration of LLMs, especially in areas of event recognition, speech processing, and industrial diagnostics. By leveraging the inherent ability to model sequential dependencies, LLMs excel at pinpointing abrupt changes in time series data, enabling robust boundary detection across diverse scenarios.

The strength of LLMs lies in their attention mechanisms, which allow to detect subtle shifts in time series data. For instance, in automatic speech recognition, LLMs improve the word and phrase segmentation accuracy by analyzing contextual dependencies in audio streams \cite{D15}. Similarly, in structural health monitoring, LLMs identify critical load-induced anomalies in bridges or buildings by isolating minute deviations in sensor data \cite{D16}. Furthermore, domain-specific fine-tuning enables LLMs to adapt to specialized contexts, enhancing the accuracy of boundary detection \cite{D17}.

To expand LLMs' utility in boundary detection, efforts should prioritize architectural innovations and domain-specific optimizations. Integrating recurrent architectures like LSTM or hybrid models could help capture long-range dependencies in time series data while reducing computational overhead  \cite{D18}. Additionally, tailoring LLMs to leverage domain-specific linguistic patterns could refine boundary precision in specialized tasks. Researchers should also explore techniques like few-shot learning or lightweight strategies variants to minimize data and resource requirements.

Overall, LLMs have redefined boundary detection in time series analysis, offering unprecedented accuracy in identifying temporal transitions across speech, infrastructure monitoring, and beyond. By addressing computational challenges and enhancing domain adaptability, future advancements will make LLM an indispensable tool for extracting actionable insights from complex and dynamic data streams, potentially driving innovation in fields such as medical diagnosis and industrial automation that rely on precise event segmentation.

\subsection{Time Series Forecasting}

Time series forecasting is crucial in fields such as economics, finance, and energy management, and has been transformed by LLMs. By leveraging the ability to discern long-term dependencies and intricate temporal patterns, LLMs outperform traditional statistical and machine learning approaches in both forecasting accuracy and adaptability \cite{D19}.

The integration of LLMs with external knowledge bases amplifies the forecasting prowess. For example, the Huawei's Pangu Weather model \cite{D21} uses historical weather data to enhance hurricane trajectory predictions. By designing a 3D Earth-Specific Transformer architecture and hierarchical time-domain aggregation, Pangu generates 24-hour global forecasts in 1.4 seconds while surpassing conventional numerical methods. Similarly in healthcare, LLMs synthesize patients' histories and clinical data to forecast disease progression, supplementing numerical predictions with natural language reports which aid clinicians in devising treatment plans \cite{D22}.

Despite their potential, LLMs face significant hurdles such as high computational costs, extensive data requirements, and prolonged training times. Moreover, privacy concerns arise when processing sensitive data like medical records. To address these, researchers propose architectural optimizations such as hybrid models integrating variational autoencoders \cite{D23} or generative adversarial networks \cite{D24}. Future work must also prioritize secure data handling protocols and improved domain adaptation directly in model training to boost relevance and reliability \cite{D26}.

\subsection{Semantic Segmentation}

Semantic segmentation in time series analysis, particularly for applications like healthcare monitoring and environmental science, involves partitioning continuous time series data into semantically meaningful segments. LLMs with advanced deep learning architectures excel at extracting high-level features from raw time series, enabling precise and context-aware segmentation.

For example, in healthcare monitoring, LLMs address limitations of the time-consuming and subjective nature of manual arrhythmia detection in traditional ECG analysis. By training on extensive ECG datasets, LLM-driven time series analysis can automatically classify arrhythmia types and detect myocardial ischemia with high accuracy \cite{D10}. Tripathi et al. \cite{D11} integrated pre-trained LLMs with electronic health records and physiological data, and provided clinicians with actionable insights for personalized treatment plans.

The LLM-driven semantic segmentation is reshaping time series analysis and offering transformative solutions. Future research can explore the specific application scenarios of LLM-driven time series analysis for different data types and task requirements. 

\section{Open issues and challenges}

Although LLM-driven time series analysis has shown significant advantages in areas of weather forecasting, industrial IoT monitoring, physiological signal analysis, etc., they still face significant challenges in processing high-dimensional, dynamic and complex time series data.

\subsection{Computational Efficiency}

Time series data usually contain tens of thousands to millions of time steps, and the computational complexity of the attention mechanism of LLMs grows quadratically with the length of the sequence, resulting in high training and inference costs. Although the sparse attention can alleviate the problem, it may lose long-range dependency information and affect prediction accuracy. In the future, it is necessary to explore dynamic sparse mechanisms, hierarchical modeling, or combine alternative architectures to balance efficiency and long-range modeling capabilities.

\subsection{Privacy and Security}

Time series data often contains sensitive information, and the data-driven nature of LLMs may lead to privacy risks. For example, models can easily expose individual health information by reversing the training data. The potential work needs to combine differential privacy, federated learning, encrypted reasoning technology, or adversarial sample detection mechanisms to ensure the two-way security of data and models. It is necessary to explore the balance between privacy protection and model performance, and establish security assessment standards for time series scenarios.

\subsection{Model Interpretability}

Although LLMs perform well in time series tasks, the ``black box'' nature restricts the interpretability in practical applications. For example, in medical diagnosis, the model needs to explain to the doctor why a specific ECG is classified as an abnormal case \cite{D10}; in financial scenarios, logical basis for detecting abnormal transactions needs to be clarified. In the future, attention weight visualization techniques or natural language interpretation generation can be combined to improve the transparency and interpretability of models.

\subsection{Real-Time Processing Capability}

Deploying LLMs for time series analysis in latency-sensitive applications requires fast inference speed, which conflicts with the inherent computational complexity of LLMs. For example, in the industrial IoT, detecting faults from high-frequency vibration signals requires continuous streaming analysis, but LLMs may encounter memory overhead and sequence dependency bottlenecks. Although model lightweighting techniques such as quantization and pruning can reduce computational load, some key transient features in time series data are discarded. Future solutions may integrate hardware algorithm collaborative design or layered streaming architecture, prioritizing fast decision-making while delaying low-priority computations.

\subsection{Data Scarcity}

Due to high annotation costs or rare events, many time series domains face a severe shortage of labeled data. For example, training LLMs to predict neonatal epileptic seizures requires annotated EEG datasets, which are both ethically and logically challenging. Current models often overfit limited samples and cannot be generalized to scenarios with insufficient representativeness. To address this issue, a hybrid approach that combines synthetic data generation (such as diffusion models based on physical laws) with meta-learning frameworks can enhance few shot generalization. In addition, an active learning pipeline that iteratively queries high-value annotations from human experts can optimize label efficiency in industrial or medical applications.

\section{Conclusions}

Time series analysis holds a signicant role in various areas, and the emergence of LLMs provides new possibilities for this field. This paper conducts a systematic study on LLM-driven time series analysis. Firstly, it establishes an evolution roadmap of AI-driven time series analysis, from the early machine learning era, to the LLM-driven paradigm, and towards establishing native temporal foundation models. Secondly, it introduces LLM-driven time series analysis techniques and organizes a technical landscape according to the workflow perspective, namely, techniques at the LLMs' input, optimization, and lightweight stages. Ultimately, possible application scenarios of LLM-driven time series analysis are highlighted, and open challenges are envisioned for further exploration. The work provides an overall introduction to LLM-driven time series analysis, and will highlight future research opportunities to advance time series analysis.

\small
\bibliographystyle{IEEEtran}
\bibliography{myreference}

\vfill

\end{document}